\documentclass[11pt]{article} 

\pdfoutput=1
\usepackage{authblk}
\usepackage{cmap}
\usepackage[T1]{fontenc}

\usepackage{graphicx}
\usepackage{amsmath}   
\usepackage{mathabx}   

\usepackage[ngerman,english]{babel}
\addto\extrasenglish{\languageshorthands{ngerman}\useshorthands{"}}

\usepackage[%
rm={oldstyle=false,proportional=true},%
sf={oldstyle=false,proportional=true},%
tt={oldstyle=false,proportional=true,variable=true},%
qt=false%
]{cfr-lm}
\usepackage[math]{blindtext}

\usepackage{cite}

\usepackage{paralist}

\usepackage{csquotes}

\usepackage{microtype}

\usepackage{url}
\makeatletter
\g@addto@macro{\UrlBreaks}{\UrlOrds}
\makeatother

\usepackage{xcolor}

\usepackage{xspace}

\DeclareFontFamily{U}{MnSymbolC}{}
\DeclareSymbolFont{MnSyC}{U}{MnSymbolC}{m}{n}
\DeclareFontShape{U}{MnSymbolC}{m}{n}{
    <-6>  MnSymbolC5
   <6-7>  MnSymbolC6
   <7-8>  MnSymbolC7
   <8-9>  MnSymbolC8
   <9-10> MnSymbolC9
  <10-12> MnSymbolC10
  <12->   MnSymbolC12%
}{}
\DeclareMathSymbol{\powerset}{\mathord}{MnSyC}{180}

\begin{document}

\title{Combining Deep Universal Features, Semantic Attributes, and Hierarchical Classification for Zero-Shot Learning}

\author[1]{Jared Markowitz}
\author[1]{Aurora C. Schmidt }
\author[1]{Philippe M. Burlina}
\author[1]{I-Jeng Wang}
\affil[1]{Johns Hopkins University Applied Physics Laboratory}
\date{}
%

\maketitle

\begin{abstract}
We address zero-shot (ZS) learning, building upon prior work in hierarchical classification  by combining it with approaches based on semantic attribute estimation.  For both non-novel and novel image classes we compare multiple formulations of the problem, starting with deep universal features in each case.  We investigate the effect of using different posterior probabilities as inputs to the hierarchical classifier, comparing the performances of posteriors derived from distances to SVM classifier boundaries with those of posteriors based on semantic attribute estimation.  Using a dataset consisting of 150 object classes from the ImageNet ILSVRC2012 data set, we find that the hierarchical classification method that maximizes expected reward for non-novel classes differs from the method that maximizes expected reward for novel classes.  We also show that using input posteriors based on semantic attributes improves the expected reward for novel classes.
\end{abstract}


\section{Introduction}\label{sec:intro}
Recent advances in deep learning have enabled significant improvements in image recognition, with some methods approaching or even exceeding the performance of humans.  However these results are predicated on the availability of large numbers of training examples of each class being considered, a requirement that is sometimes difficult to fulfill.  Attention has therefore shifted to the more realistic situation where the number of training examples per class is more variable, with plentiful examples of some classes but few to no examples of others.  The task of classifying samples from classes with no training examples is known as \textit{zero-shot learning}. This problem has interest from many practical standpoints.  Among them are the automatic labeling of novel data sets (avoiding the burden of classifier retraining), lifelong learning in open sets/universes, decision-making of autonomous agents, and the study of how humans perform the task~\cite{biederman1987recognition, fei2006one}. 
 
In this work we investigate combinations of two different yet complementary approaches to zero-shot learning: approaches based on hierarchies and approaches based on semantic attributes.  One may use hierarchical classification to characterize objects by trading specificity for accuracy.  The thought is that while a novel object cannot be classified exactly it may still be placed in a more general category, possibly providing useful and actionable information.  To illustrate this consider a classifier that, while having never before seen an apple, can still recognize an apple as a fruit.  The information thus provided allows one to take appropriate action (i.e. to safely eat the object). 

In contrast, approaches based on semantic attributes exploit the capability of humans to categorize yet unseen classes by learning mappings from images to human recognizable attributes.  One assumes that while training data for novel classes do not exist, knowledge regarding the semantic qualities of the unknown classes is available.  This information may come from human annotation or may be generated using other mechanisms (for instance extraction from a corpus of documents).  To illustrate the utility of this approach, consider a scenario where tigers are not included in a data set but are known to be large, striped, and carnivorous.  From this we may be able to directly classify a tiger or to at least infer correct actionable information about it (i.e. to steer clear of it!).  The semantic attributes provide an intermediate mapping between object data and class labels that is intended to enable sufficiently accurate classification of even novel objects.

Hierarchical and attribute-based classification methods provide different types of output information; the former gives a predicted position in an established hierarchy while the latter provides a ranked listing of the classes and their estimated probabilities.  Rank may be used as a means of comparing these approaches if, for the hierarchical classifiers, it is generated by counting the leaf nodes below the predicted position in the hierarchy.  In this study we find that, when allowing for novel classes, the rank of the true class obtained from an attribute-based classifier is generally smaller (better) than the number of potential classes remaining after hierarchical classification.  However the former gives no notion of hierarchical context; in fact the top-ranked classes may come from very different regions of the ground truth hierarchy. 
 
Here we seek to leverage the complementary strengths of these approaches by exploring methods for combining them.
We work in both directions.  Starting with attribute-based classification, we use the resulting class posteriors as the input to hierarchical classifiers in order to produce improved average information gain in zero-shot applications.  
Conversely, we show that the results of hierarchical classification can be used to pare down the ranked lists produced by attribute-based analyses. 

Throughout we consider only zero-shot classification problems where it is not known a-priori whether a given test image comes from a novel class.  The corresponding classification task has at different times been referred to as ``blind,'' ``uninformed,'' or ``generalized''~\cite{xian2017zero} zero-shot learning.  This scenario is clearly both more likely to be encountered in real life and more challenging than ``informed'' zero shot analysis, where one knows beforehand whether or not an object is novel. 

\section{Related Work}
A considerable amount of effort has recently been dedicated to zero-shot learning, and this effort can be categorized using various taxonomies. One such taxonomy is based on the output of the method; either a position in a class hierarchy~\cite{DengKrauseBergFei-Fei_CVPR2012,MVA2017} or a class ID (most methods, including \cite{lampert2014attribute, norouzi2013zero,burlina2015zero}). Another breakdown could be based on the problem considered; either informed ZS (most studies thus far) or uninformed ZS~\cite{xian2017zero,burlina2015zero}.  Finally one could distinguish efforts based on the type of manifold embedding used; many use semantic attributes (most work in ZS, e.g. \cite{lampert2014attribute}) while some use a mixture of existing classes~\cite{norouzi2013zero} or a compatibility in some nonlinear manifold  (see examples in ~\cite{xian2017zero}).

The most common zero-shot strategy to date has been to learn an intermediate mapping from automatically computed features to semantic attributes before classification.  As mentioned above, the hope is that the identified attributes will enable the classification of even novel objects~\cite{palatucci2009zero, yu2010attribute, lampert2014attribute, norouzi2013zero, hoo2014zero, frome2013devise, JayaramanG14, burlina2015zero, chao2016, disney16}.  Another strategy was described in \cite{fei2006one}, where a Bayesian framework was used to learn a probabilistic representation for each class that was improved upon with each example.  Finally two other notable approaches~\cite{hoo2014zero, salakhutdinov2013learning} represent objects using a topic model. In particular \cite{salakhutdinov2013learning} introduces ``hierarchical deep'' learning, wherein an image is first input to a deep Boltzmann machine and subsequently passed to a nonparametric Bayesian topic model.  The two stages are trained together, with the nonparametric topic model enabling the identification of new classes at different levels of granularity.  

 This report builds upon the results of ~\cite{MVA2017}.  It uses class rankings and attribute-based posterior probabilities derived from the direct and indirect methods described in \cite{burlina2015zero}, incorporating them into hierarchical  classification schemes. Numerous approaches to cost-sensitive and hierarchical classification have previously been investigated, including those described in ~\cite{dekal2004, cesa2006, zweig2007, marszalek2007, marszalek2008, bengio2010, Deng10000, Deng2011, gao2011, hwang2011, zhao2011, DengKrauseBergFei-Fei_CVPR2012}. 
Here we focus on the formulation in \cite{DengKrauseBergFei-Fei_CVPR2012}, which was among the first to determine object levels of abstraction by optimizing the trade-off between accuracy and specificity. We more thoroughly evaluate the performance of the Dual Accuracy Reward Trade-off Search (DARTS) and Maximum Expected Reward (MAX-EXP) approaches on novel classes than \cite{DengKrauseBergFei-Fei_CVPR2012} and extend both methods to incorporate information based on semantic attributes, quantifying the impact.

One previous work that uses semantic attribute estimation in conjunction with a hierarchical approach is \cite{howtoxfer}.  Therein the differing meanings of attributes in different parts of a hierarchy are considered.  Sets of both distinguishing attributes and refined attribute classifiers are determined at each inner node, allowing improved zero-shot classification.

\section{Zero-Shot Classification Methods}
We now formally describe the hierarchical and attribute-based classification methods employed in our analysis.

\subsection{Hierarchical Classifiers: DARTS and MAX-EXP}
Hierarchical classification approaches trade off accuracy and specificity to generate appropriate classifications for objects of varying certainties.  As mentioned above, the DARTS algorithm from \cite{DengKrauseBergFei-Fei_CVPR2012} explicitly optimizes this trade-off.  DARTS maximizes the information gained about an image (quantified by the reduction in potential leaf classes) subject to a minimum accuracy guarantee.  The optimization is formulated as 
\begin{equation}
\begin{aligned}
& \underset{f}{\text{maximize}}
& & R(f) \\
& \text{subject to}
& & \Phi(f) \geq 1 - \epsilon,
\end{aligned}
\end{equation}
where $R(f)$ is the average reward of classifier $f$, $\Phi(f)$ is the expected likelihood that the true class is the chosen node or is one of its descendants, and $\epsilon$ is the maximum allowable error rate.  This problem is parameterized using a Lagrange multiplier via
\begin{equation}
L(f, \lambda) = R(f) + \lambda(\Phi(f) -1 + \epsilon).
\end{equation}
Here the Lagrange function $L$ depends on the classifier $f$ and multiplier $\lambda$.  Since it can be shown that $\Phi(f)$ and $R(f)$ are non-decreasing in opposite directions, $\lambda$ is determined using a binary search.  The resulting optimal classifier is 
\begin{equation}
f_\lambda(x)  = \underset{v \in \mathcal{V}}{\mathrm{argmax}}(r_v + \lambda)p_{Y|X}(v|x),
\end{equation}
where node $v$ in the hierarchy $\mathcal{V}$ has reward $r_v$ and the posterior probability of image $x$ being at or a descendant of node $v$ is $p_{Y|X}(v|x)$.  This posterior is obtained by training one-vs-all linear SVM classifiers on the set $Y$ of leaf node classes of the hierarchy, using Platt scaling~\cite{platt1999probabilistic} to generate a probability distribution over $Y,$ and determining inner node probabilities by summing up the tree.


Several additional hierarchical classifiers are mentioned in \cite{DengKrauseBergFei-Fei_CVPR2012}.  Most relevant to our investigation is the MAX-EXP classifier, which provides a more direct approach to solving (1).  MAX-EXP chooses the node with highest expected reward subject to a threshold in posterior probability which is chosen to provide a maximum error rate of $\epsilon$ in the validation data.  MAX-EXP can be written as 
\begin{equation}
 f(x)  = \underset{v \in \mathcal{V} \mathrm{ : } p_{Y|X}(v|x) > \theta}{\mathrm{argmax}} r_vp_{Y|X}(v|x),
\end{equation}
where $\theta$ is the posterior probability threshold. 

Finding appropriate values for the $\lambda$ and $\theta$ parameters that govern the DARTS and MAX-EXP classifiers can be difficult, particularly when high accuracies are desired. To see this, consider first $\lambda$.  The upper bound on its value is determined by the smallest ratio of true class posterior to any other posterior in the training data.  This ratio can be arbitrarily small when the posterior probability is flawed, leading to an arbitrarily large $\lambda$ to compensate.  Large $\lambda$ values are problematic because they push all instances toward the root node, where no information is gained.  A similar behavior occurs in MAX-EXP, where high accuracy requirements force $\theta$ to be pushed arbitrarily close to 1.  This again results in uninformative classifications.  Hence both methods must be tuned with care; accuracy requirements must be high enough to ensure that the classifier generally chooses the right branch but not so high that the classifier fails to proceed down the branch.

Another issue with both DARTS and MAX-EXP arises because of discrepancies in the difficulty of classification decisions in different parts of the hierarchy.  Since only one overall accuracy is specified, both algorithms will sometimes neglect more difficult classes while maximizing their performance on easier classes.  Though we do not address it further here, this issue may be minimized by adjusting the rewards to focus attention on different branches of the tree.

\subsection{Classification via Semantic Attributes}

As mentioned above, semantic attributes can be used as an intermediate step between features automatically generated from images and classification.  This step provides some notion of human intuition and can be applied to both known and novel classes, enabling zero-shot learning.  Here we apply the direct and indirect approaches for attribute-based classification formulated in \cite{burlina2015zero, lampert2009learning}.  

To set up the problem, call the set of image features $\mathcal{X}$, the set of non-novel labels $\mathcal{Y}$, and the set of novel labels $\mathcal{Z}$.  We start with a set of labeled examples $\left( {\bf x}_1, l_1 \right) ... \left({\bf x}_n, l_n \right) \in \mathcal{X} \bigtimes \mathcal{Y}$, where ${\bf x}_1,\ldots,{\bf x}_n$ are image features obtained using a deep convolutional neural network (Section 4).  Also known a priori is the class to attribute matrix $\bf{V}$, which gives ground truth attribute values for both non-novel and novel classes.  As in \cite{burlina2015zero} ternary values are used for each attribute, with the possible values of 1, 0, and -1 denoting ``yes,''
 ``no,''
  and
   ``indeterminate'' respectively.  The ``indeterminate'' response indicates an attribute categorization that is either unclear or is not applicable to the image in question. 

In the direct approach, each attribute is learned using a classifier trained using data from $\mathcal{X}$.  The class-to-attribute matrix is used to create a set of attribute-labeled training examples $\left( {\bf x}_1, u_1\right)$, ...$\left({\bf x}_n, u_n\right) \in \mathcal{X} \bigtimes \left\{-1, 0, +1\right\}$ for each attribute $j=1,..,N_a$.  $N_{a}$ attribute classifier models are built from these data using Linear Support Vector Machines (LSVM) and are used to infer the feature to attribute mapping ${\bf {\tilde v}}({\bf x})$. 

In the indirect method, the posterior distribution on class given features $P(Y|\bf X)$ is first learned using one-vs-all linear SVM classification followed by Platt sigmoid scaling \cite{platt1999probabilistic}.  This posterior is used to obtain attribute estimates through a weighted average:
\begin{equation}
E_P [v_j] = \sum_{ i \in Y}   P(\mathrm{class}=i|{\bf X} ={\bf x}) v_{i,j}.
\end{equation} 
The result is thresholded to produce a final attribute estimate in $\left\{-1, 0, +1\right\}$.

Using the inferred attributes, class probability distributions (over both non-novel and novel classes) are obtained using the maximum likelihood (ML) method in ~\cite{burlina2015zero}.  The approach computes class posterior likelihoods given the estimated attributes using the measured error rates of each attribute classifier in a validation data set and assuming independence of errors.  For inferred attribute vector $\bf v$, the likelihood of class $i$ is determined using 
\begin{equation}
P({\bf x}|y_i)  \sim  P(\widehat{\bf v} ={\bf v}_i) \approx  \prod_{j=1,..,N_a} P(\widehat{v}(j)|v_{i,j}), 
\label{e: prob_attr}
\end{equation}
where the ${\bf v}_i$ are the attribute vectors in ${\bf V}$ corresponding to the class ${i \in \mathcal{A} = \mathcal{Y} \cup \mathcal{Z}}$.
This ML-based class ranking was shown in previous experiments to perform similarly to or slightly better than the distance-based ranking in \cite{burlina2015zero}.  It was chosen here because unlike distance-based methods it immediately provides class posterior distribution estimates that can be used as inputs to hierarchical classification algorithms. 

\subsection{Combination Approaches}
There are at least two ways that attribute-based classification can be integrated into a hierarchical approach.  The first is to use the class posterior probability estimates derived from the attribute analysis as the inputs to the hierarchical classifier.  These posteriors provide additional semantic information to the hierarchical classifier and in particular may enable improved performance on novel classes.  The second way of combining approaches is to use the hierarchical classification result (with standard prior) to choose the number $N$ of leaf nodes to consider from the class rankings provided by the attribute-based approach.  As with the first combination approach, this ``TOPN'' method seeks to improve the average information gained for images from both non-novel and novel classes.

\begin{figure*}[t]
  \centering
  \includegraphics[width=0.88\linewidth]{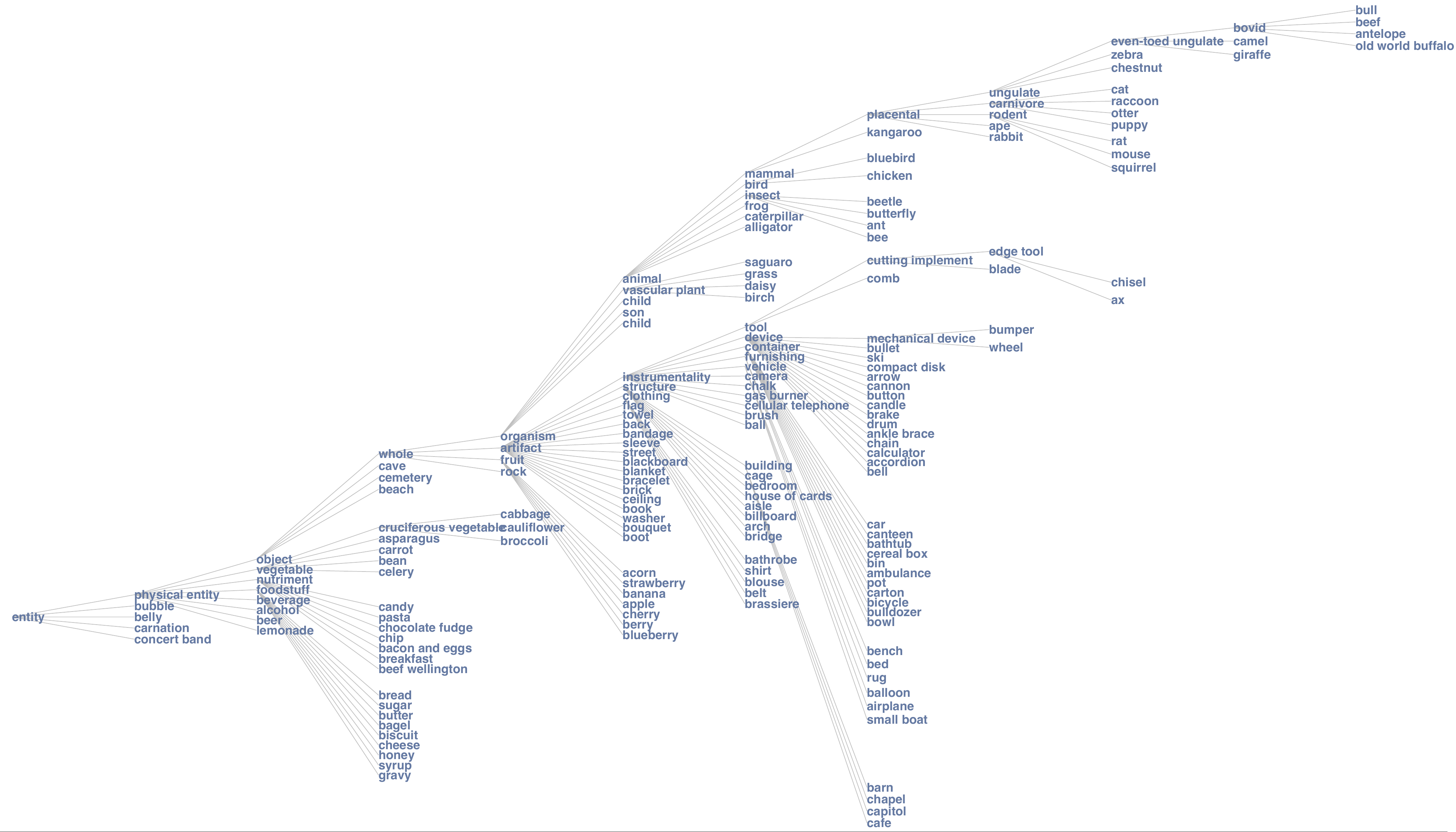}
  \caption{Attribute-based hierarchy of 150 ImageNet Classes.}
  \label{f:hierarchy}
\end{figure*}

\section{Experiments and Results}

\subsection{Data Processing}

This analysis was conducted using a data set comprised of 192,870 images from 150 ImageNet \cite{deng2009imagenet} classes.  These classes were deliberately chosen to be diverse, spanning groups from animals to household objects (Figure 1).  Thirty of the classes were withheld from training for zero shot analysis.  Images from the remaining 120 classes were split into training, validation, and testing sets according to a $90\%-5\%-5\%$ split.  Each image was processed using the OVERFEAT deep convolutional network \cite{sermanet2013overfeat}, with the output of the last fully connected layer (fc7) being taken as the (4096-dimensional) feature vector used in further analysis.

In order to generate semantic information, a set of 218 questions was answered by a group of humans about each class\cite{palatucci2009zero}.  These straightforward questions included queries such as ``is it man-made?'' 
and
 ``is it edible?''  For each class, the answers to each question were averaged and thresholded to give values of Yes (1), Maybe (0), or No (-1).  This resulted in a 218-dimensional attribute vector being associated with each class.  

The ground truth hierarchies required for application of hierarchical classification algorithms were generated by clustering these attribute vectors.  Specifically, correlation distances were computed between each pair of attribute vectors and clustered to determine higher level groupings\cite{Mullner2011}.  The resulting inner nodes were then identified by cross-referencing with WordNet~\cite{WordNet}; a given inner node was matched to the nearest common parent of all its children in the WordNet tree.  The ground truth hierarchy used for the following experiments and generated in this fashion is shown in Figure 1.

As described above, the hierarchical classification algorithms we address require posterior class probability estimates as input.  These distributions were generated using the following approach.  The image feature vectors were first used to train a one-vs.-all SVM classifier for each training class.  Here the LIBLINEAR default C=1 \cite{fan2008liblinear} was used instead of the C=100 values used in \cite{DengKrauseBergFei-Fei_CVPR2012}. For each input image, Platt scaling \cite{platt1999probabilistic} was then applied to generate a probability estimate for each leaf node.  Higher level node probabilities were inferred through summation of the probabilities of their descendant leaf nodes, moving up the hierarchy.  Note that the hierarchy was pruned to remove the 30 novel classes in this procedure.

\subsection{Comparison of Hierarchical and Attribute-based Methods}
Using this data set, we first compared the abilities of hierarchical and attribute-based methods to reduce the number of potential classes.  We used two metrics: average information gain and mean rank accuracy (MRA). These metrics required slightly different interpretations for the two methods due to the differences of their outputs. 

For hierarchical classifiers, the information gain $r(v)$ used was the reward used in ~\cite{DengKrauseBergFei-Fei_CVPR2012}: 
\begin{equation}
r(v) = \mathrm{log}_2\left|\mathcal{Y}\right| - \mathrm{log}_2 \sum_{y \in \mathcal{Y}}\left[v \in \pi(y) \right].
\label{e: reward}
\end{equation}
Again $v$ represents the chosen node and $\mathcal{Y}$ the set of leaf nodes. The set of ancestors of a leaf node $y$ is $\pi(y)$.  For the attribute-based methods, the information gain was derived from the rank of the true class $R_{true}$ provided by the method:
\begin{equation}
r = \mathrm{log}_2\left|\mathcal{Y}\right| - \mathrm{log}_2 \left( R_{true} \right).
\end{equation}
Both of these quantities were averaged over the testing data.

For the flat attribute-based classifiers, the mean rank accuracy was defined as  
\begin{equation}
\mathrm{MRA} = 1 - \frac{\mathrm{mean}(R_{true})}{|\mathcal{Y}|}.
\end{equation}
For the hierarchical methods, contributions to the MRA took different forms depending on whether or not the classification was on the right branch.  Thinking in terms of a ranked list of leaf nodes, all classifications brought the descendants of the chosen node to the front of the list.  Correct classifications contributed a rank of half of the number of leaf node descendants of the chosen node (with a minimum contribution of 1).  Incorrect classifications contributed a rank of the sum of the number of descendants of the chosen node plus half of the number of remaining leaf classes.


Table 1 shows the values for these metrics for both direct and indirect attribute-based methods as well as the DARTS and MAX-EXP hierarchical classifiers.  Note that both DARTS and MAX-EXP were tuned to maximize the mean rank accuracy of each experiment, resulting in average information gains that were slightly sub-optimal.  As expected, the flat attribute-based classifiers outperform the hierarchical methods in each case.  However it should be remembered that the superior rank accuracy of the attribute-based methods does not come with any indication of where the true class is in the tree, nor does it include hierarchical context.  The top few classes ranked by the attribute analysis may come from widely separated regions of the hierarchy, casting doubt on the true nature of the object.  In contrast a hierarchical placement may include more leaf classes than its counterpart, but these leaf classes will all have at least some common ancestry.  Hence the decrease in rank accuracy observed from attribute-based methods to hierarchical classifiers may be viewed as the cost of incorporating hierarchical context.

\begin{table}
\centering
\begin{tabular}{ | c | c | c | c |}
\hline
Algorithm & Data Set & Avg. Inf. Gain & MRA \\
\hline
\hline
Direct MLE & Non-novel & 6.105 & 0.9507 \\
Indirect MLE & Non-novel & 6.072 & 0.9338 \\
DARTS & Non-novel &  4.9961 & 0.8636 \\
MAX-EXP & Non-novel & 5.0901 & 0.8587 \\
\hline
\hline
Direct MLE & Novel & 3.445 & 0.8241 \\
Indirect MLE & Novel & 3.416 & 0.8016 \\
DARTS & Novel & 0.4374 & 0.5655 \\
MAX-EXP & Novel &  0.7170 & 0.5952 \\ 
\hline
\end{tabular}
\caption{Comparison of attribute-based and hierarchical classification methods.  Average information gain and mean rank accuracy (MRA) are shown.} 
\end{table}

\subsection{Comparison of Hierarchical Approaches}

Several variations of the hierarchical classification strategies described in Section 3 were evaluated.  Both DARTS and MAX-EXP were tested using input posterior probabilities generated by standard 1-vs-all SVM classification followed by Platt scaling as well as posterior probabilities generated by the direct and indirect attribute analyses \eqref{e: prob_attr}.  We further quantified the impact of the ``TOPN'' cross-referencing approach described above.  Note that the branches of the ground truth hierarchy that led to novel classes were pruned when using conventional posteriors but not when using attribute-based posteriors.

The different methods were characterized by plotting the average reward versus the average accuracy for both non-novel and novel instances in the testing set (Figures 2 and 3).  All classification tasks were done using the ``blind'' approach to zero shot, meaning that it was not known whether or not an image came from a class contained in the training set.  Each trace on each plot was generated by considering a range of parameters for the hierarchical classifiers; $\lambda$ for DARTS and $\theta$ for MAX-EXP.

\begin{figure}
\includegraphics[width=1.0\linewidth]{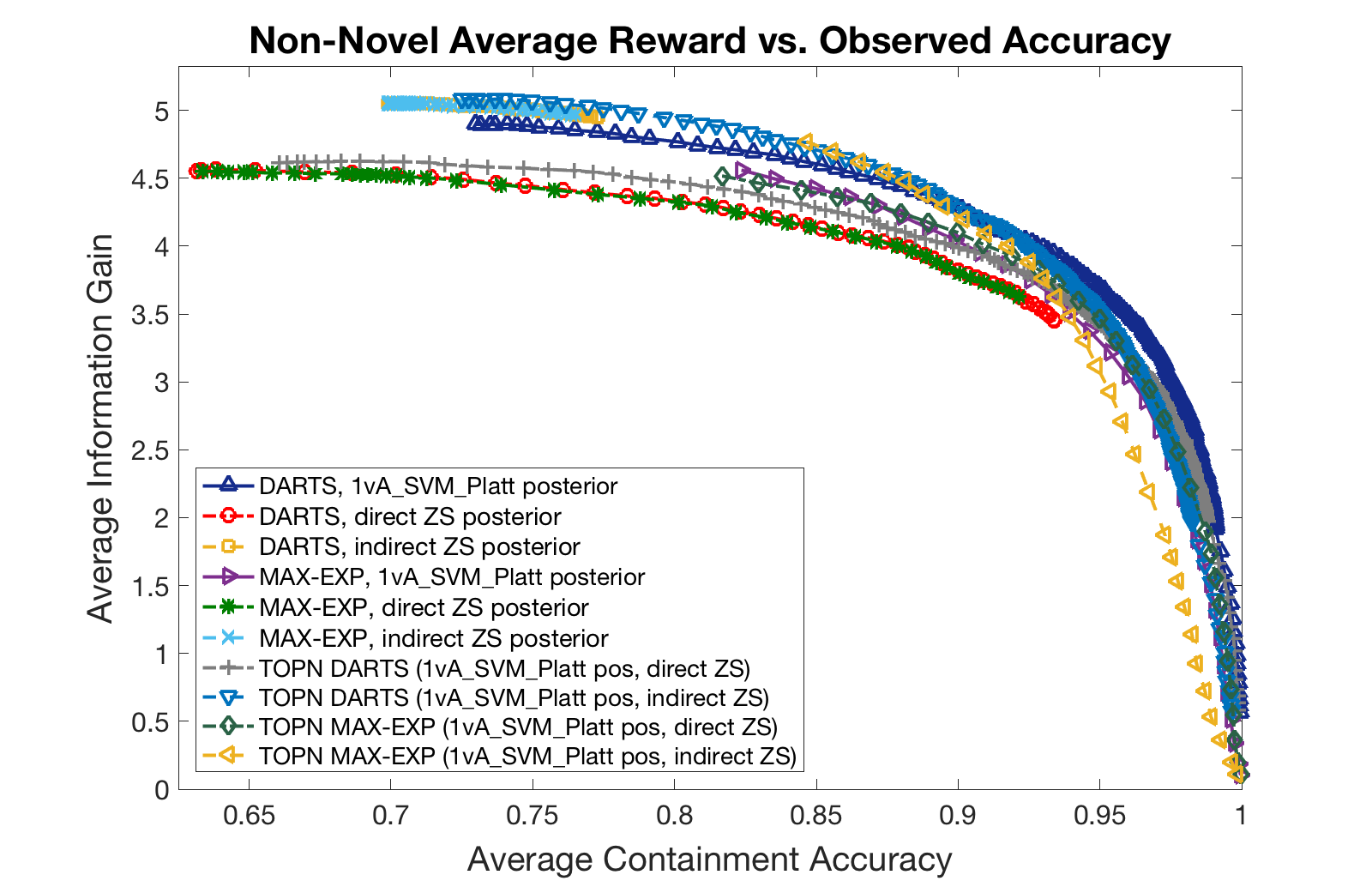}
\caption{Hierarchical classification of images from previously seen (non-novel) classes.}
\label{f:non_novel}
\end{figure}

\begin{figure}
\includegraphics[width=1.0\linewidth]{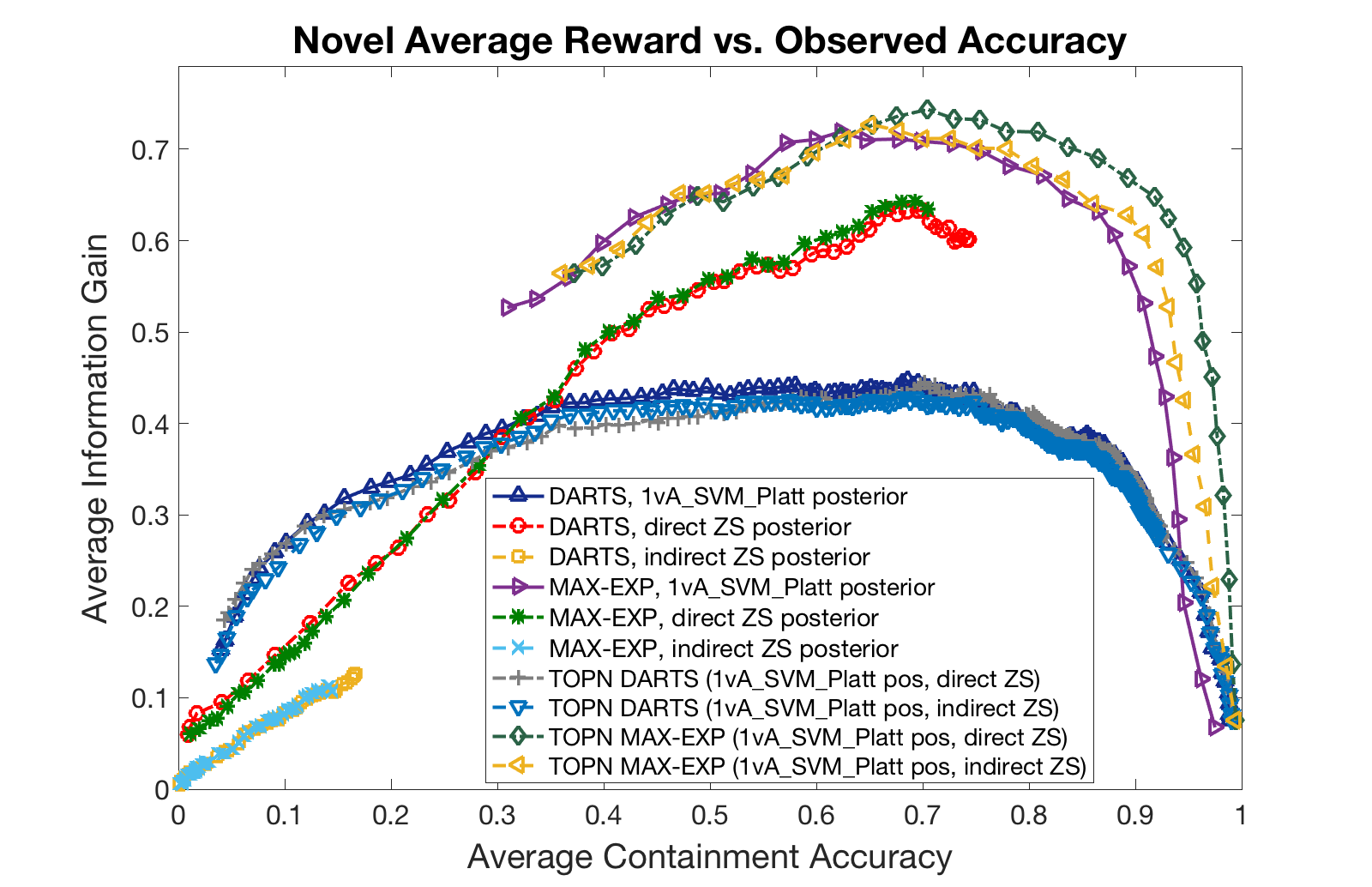}
\caption{Hierarchical classification of images from previously unseen (novel) classes.}
\label{f:novel}
\end{figure}

Several observations can be made about the resulting plots.  As expected, the average information gain on images from previously seen classes is far greater than the average information gain on images from previously unseen classes.  The variation in performance of the methods on non-novel classes is relatively small; on our logarithmic scale \eqref{e: reward} the difference between an average reward of 4.5 and 5.0 is only about 2 potential classes per image.  For the most part it appears that methods with conventional posteriors (1-vs-all SVM classifiers followed by Platt scaling) slightly outperform those with attribute-based posteriors.  Differences in the performances of the algorithms are much more pronounced over novel classes.  To see this, note that (because of the logarithmic scale) the difference between an average reward of 0.4 and an average reward of 0.7 is about 21 leaf classes.  For previously unseen classes, MAX-EXP generally outperforms DARTS.  Posteriors derived from the direct method of attribute classification are seen to provide significant improvements to the performance of DARTS for moderate accuracy levels.  The two highest-performing methods are the TOPN MAX-EXP methods, illustrating the value of incorporating both attribute-based posteriors and rankings.

\begin{figure}
\includegraphics[width=1.0\linewidth]{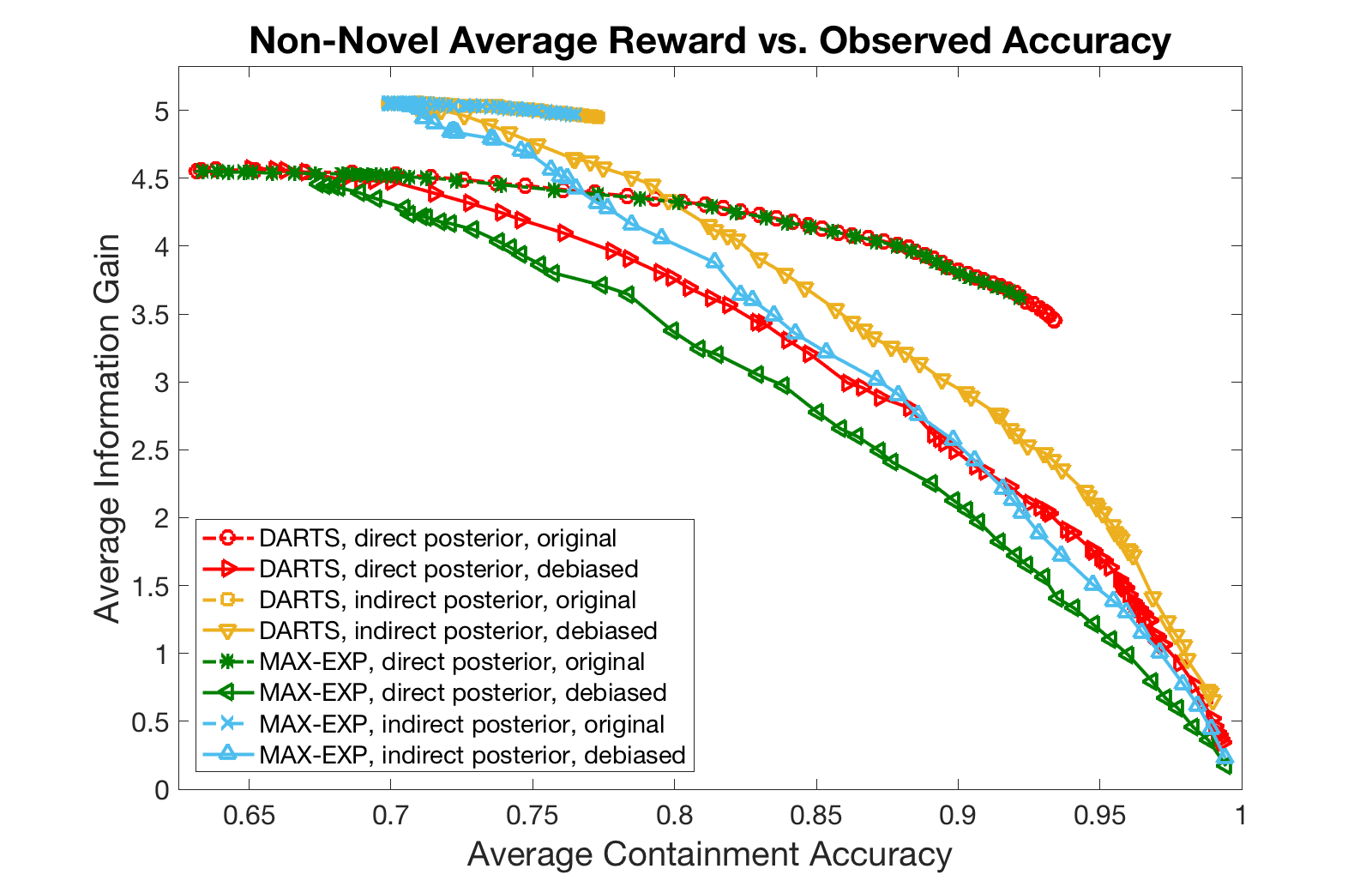}
\caption{Effect of debiasing on information gain in hierarchical classification of non-novel classes.}
\label{f:debiasing}
\end{figure}

One notable aspect of Figures 2 and 3 is the failure of some traces to reach full accuracy at any point.  This occurs mainly in methods where attribute-based posteriors are used, and reflects a bias in these posteriors.  The lack of convergence to full accuracy for even very large values of  $\{\lambda$, $\theta\}$ occurs because of instances where the posterior strongly favors an incorrect leaf node.  We were able to correct this problem in the non-novel classes by computing the confusion matrix over the validation set and using it to re-normalize the posteriors computed for the test set.  The result of this procedure is shown in Figure 4.  While the debiasing successfully allows the methods to reach full accuracy for much more reasonable (i.e. lower) tunings of $\{\lambda$, $\theta\}$, it does so while producing smaller average rewards for the highest accuracy points of the original traces.

\section{Conclusion}
We have compared and combined multiple methods for hierarchical and attribute-based classification, looking at previously seen and previously unseen classes separately.  The two approaches have complementary strengths; attribute-based methods excel at providing rankings with the true class near the top while hierarchical methods provide context, allowing inference of object characteristics.  We investigated two means of incorporating information based on semantic attributes into hierarchical analysis: through leaf node posterior probabilities and through cross-referencing with ranked class lists derived using attributes.  All flavors of both approaches yielded similar classification results (quantified in terms of average information gain) for non-novel classes while yielding more significant gains for novel classes.  We additionally showed that biasing effects due to inaccurate attribute-based posteriors may be removed through renormalization by the confusion matrix, albeit at the cost of lowering the average information gained.

\subsubsection*{Acknowledgments}
We thank the authors of \cite{palatucci2009zero} for providing the class to attribute mapping of their data set.

\bibliographystyle{splncs03}

\bibliography{new_references}
\end{document}